\documentclass{article}
\usepackage{spconf,amsmath,graphicx}

\usepackage{enumitem}
\setlist{nosep, leftmargin=14pt}

\usepackage{mwe} 

\usepackage{hyperref}
\newcommand\Tstrut{\rule{0pt}{2.0ex}}         
\newcommand\Bstrut{\rule[-0.9ex]{0pt}{0pt}}   
\usepackage{caption}
\title{Dual-Domain Cross-Iteration Squeeze-Excitation Network for Sparse Reconstruction of Brain MRI}

\name{Xiongchao Chen$^{1,2}$\sthanks{Please contact the authors at \href{mailto:xiongchao.chen@yale.edu}{xiongchao.chen@yale.edu}}, Yoshihisa Shinagawa$^{1}$, Zhigang Peng$^{1}$, Gerardo Hermosillo Valadez$^{1}$}
\address{$^{1}$ Siemens Healthineers, Malvern, PA 19355, USA \\ 
$^{2}$ Department of Biomedical Engineering, Yale University, New Haven, CT 06511, USA}

\begin{document}
%
\maketitle
\begin{abstract}
Magnetic resonance imaging (MRI) is one of the most commonly applied tests in neurology and neurosurgery. However, the utility of MRI is largely limited by its long acquisition time, which might induce many problems including patient discomfort and motion artifacts. Acquiring fewer k-space sampling is a potential solution to reducing the total scanning time. However, it can lead to severe aliasing reconstruction artifacts and thus affect the clinical diagnosis. Nowadays, deep learning has provided new insights into the sparse reconstruction of MRI. In this paper, we present a new approach to this problem that iteratively fuses the information of k-space and MRI images using novel dual Squeeze-Excitation Networks and Cross-Iteration Residual Connections. This study included 720 clinical multi-coil brain MRI cases adopted from the open-source deidentified fastMRI Dataset \cite{zbontar2018fastmri}. 8-folder downsampling rate was applied to generate the sparse k-space. Results showed that the average reconstruction error over 120 testing cases by our proposed method was $2.28 \pm 0.57\%$, which outperformed the existing image-domain prediction ($6.03 \pm 1.31\%$, $p < 0.001$), k-space synthesis ($6.12 \pm 1.66\%$, $p < 0.001$), and dual-domain feature fusion ($4.05 \pm 0.88\%$, $p < 0.001$).
\end{abstract}
\begin{keywords}
Dual-domain deep learning, cross-iteration residual connection, squeeze-excitation network, sparse MRI reconstruction, multi-coil parallel imaging
\end{keywords}
\section{Introduction}
\label{sec:intro}
Magnetic resonance imaging (MRI) has rapidly become an essential clinical diagnosis and management tool of neurology \cite{arnold2002mri}. MRI can discriminate different anatomical structures with extreme high resolution and contrast, which makes MRI essential for clinical diagnosis of a wide variety of disorders including neurological and oncological diseases. 

\begin{figure}[htb!]
\centering
\includegraphics[width=0.44\textwidth]{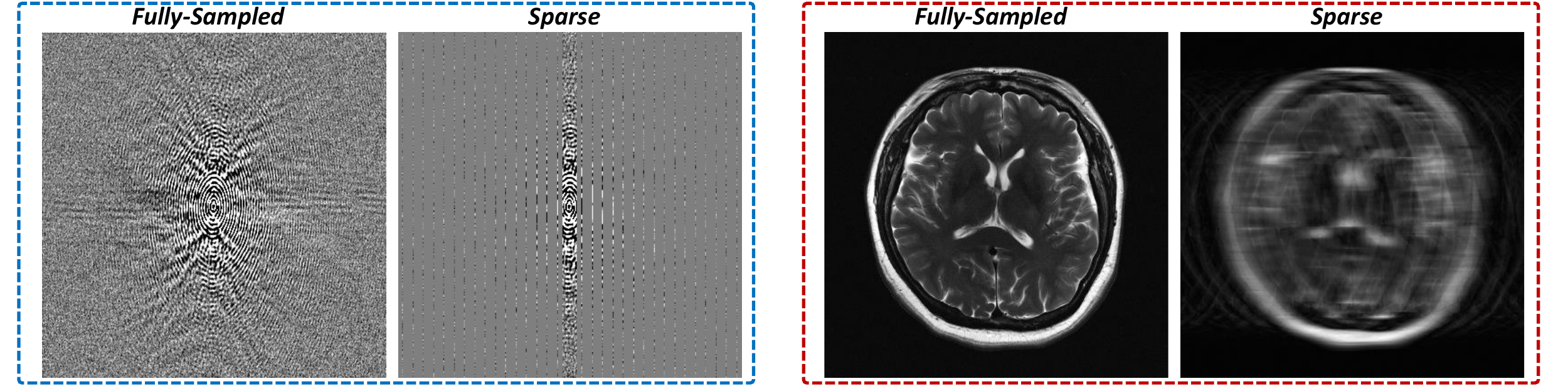}
\caption{Fully-sampled and sparse k-spaces and images.}
\label{fig:intro}
\vspace{-0.5cm}
\end{figure}

However, the long scanning time of MRI might induce many problems including patient discomfort, high exam cost, motion artifacts, and low patient throughput. Thus, reducing the MRI scanning time is necessary for accurate and efficient MRI diagnosis. One approach commonly used in current clinical practice for MRI acceleration is Parallel Imaging \cite{griswold2002generalized}. It utilizes multiple receiver coils to simultaneously acquire the multi-coil information. The other potential approach is downsampling the k-space measurements. However, the reconstructed images from the downsampled sparse k-space will display severe aliasing artifacts as shown in Fig.~\ref{fig:intro}, which largely reduce the accuracy of clinical diagnosis.

Deep learning has shown promising results in sparse reconstruction of MRI. Existing deep learning methods can be generally classified into three categories. The first category applied the sparsely reconstructed MRI images as input of neural networks to predict the synthetic fully reconstructed images \cite{jethi2020dual}. The second category utilized the sparse k-space as input of networks to generate the synthetic full-view k-space \cite{du2020multiple}. The third category combines the features of k-space and images in a dual-domain manner by Fourier Transform, to restore the full-view k-space \cite{eo2018kiki}. However, the cross-iteration features were ignored in previous dual-domain methods. In order to incorporate the cross-iteration information, we present a novel Dual-Domain Cross-Iteration Squeeze-Excitation Network (DD-CSENet). Our contribution are: 1) we develop and incorporate the dual Squeeze-Excitation Network (SENet) into the dual-domain framework to improve the reconstruction accuracy; 2) we propose the Cross-Iteration Residual Connection structure to fuse the information across different iterations to further improve the network performance.

\begin{figure*}[htb!]
\centering
\includegraphics[width=0.90\textwidth]{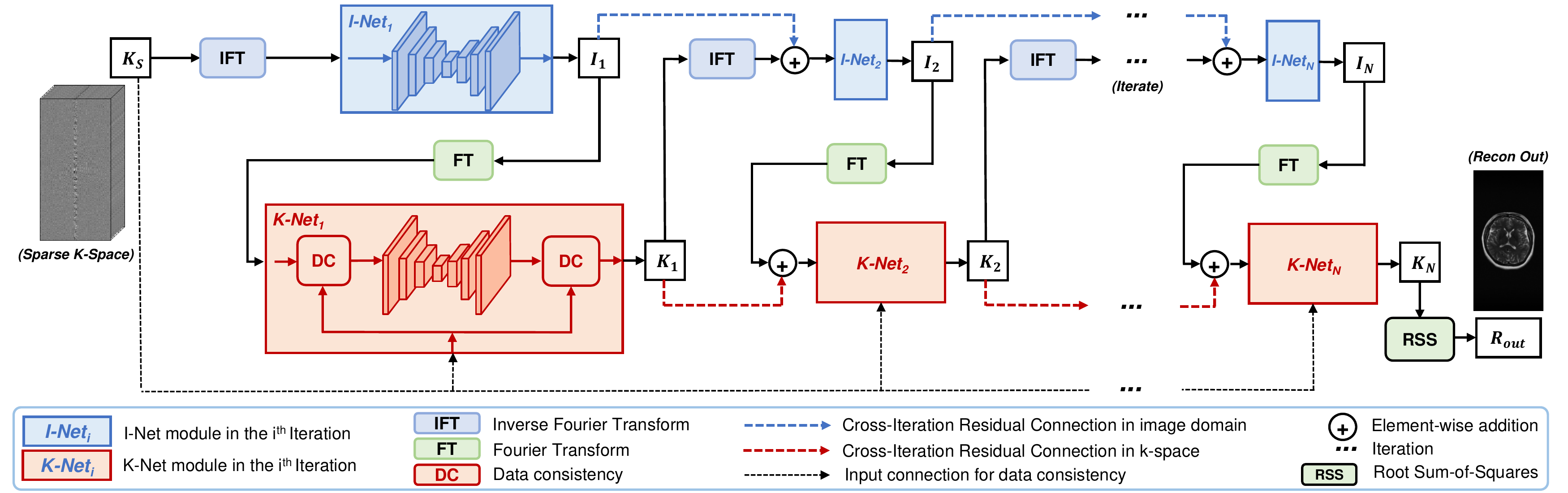}
\caption{The architecture of DD-CSENet. Each \textit{I-Net} or \textit{K-Net} module contains a dual Squeeze-Excitation Network (SENet). Each \textit{K-Net} contains two additional data consistency (DC) modules. The Cross-Iteration Residual Connections maintain and link the $I_1$ and $K_1$ to the input of \textit{I-Net$_2$} and \textit{K-Net$_2$} of the next iteration for information retaining. After $N$ iterations, $K_N$ was reconstructed into $R_{out}$.}
\label{fig:diagram}
\vspace{-0.45cm}
\end{figure*}

\section{Materials and Methods}
\label{sec:method}
\subsection{Dataset preprocessing}
In this study, 720 clinical 16-coil T2-weighted brain MRI slices were adopted from the fastMRI Dataset \cite{zbontar2018fastmri}. The downsampled sparse k-space was generated by masking the fully sampled k-space lines in the phase encoding direction. The central 4$\%$ regions of the full k-space were remained, and the average downsampling rate was set as 8 in this study. 

Root Sum-of-Square (RSS) algorithm was applied in this study for the multi-coil MRI reconstruction, formulated as:
\begin{equation}
    M_{RSS}=\left( \sum_{i=0}^{n_c}{\left| \hat{M}_i \right|^2} \right) ^{\frac{1}{2}},
\end{equation}
where $n_c$ is the number of coils and $\hat{M}_i$ is the image of the $i^{th}$ coil. In total, 480, 120, 120 cases were used for training, validation, and testing, respectively.

\subsection{Dual-Domain Cross-Iteration Framework}
The architecture of the proposed DD-CSENet was presented in Fig.~\ref{fig:diagram}. The sparse k-space $K_S$ is input to the \textit{I-Net$_1$} module after reconstruction, to generate the output $I_1$:
\begin{equation}
    I_1 = \mathcal{H}_{I_1}(\mathcal{F}^{-1}(K_S)),
\end{equation}
where $\mathcal{F}^{-1}$ is the inverse Fourier Transform (IFT) for reconstruction. $\mathcal{H}_{I_1}$ refers to the dual Squeeze-Excitation Network (SENet) in \textit{I-Net}$_1$. The structure of SENet will be introduced later in section 2.3.

Then, $I_1$ was input to the \textit{K-Net$_1$} module after image projection, to generate the output $K_1$:
\begin{equation}
    K_1 = \mathcal{P}(\mathcal{H}_{K_1}(\mathcal{P}(\mathcal{F}(I_1), K_S)), K_S),
\end{equation}
where $\mathcal{F}$ is the Fourier Transform (FT) for projection from image to k-space. $\mathcal{H}_{K_1}$ denotes the SENet in the \textit{K-Net}$_1$ module. $\mathcal{P}$ is the Data Consistency (DC) module. The output $K_{DC}$ of DC with $K_{pre}$ and $K_S$ as inputs is formulated as:
\begin{equation}
    K_{DC}(x) = 
    \begin{cases}
      \frac{\lambda K_{pre}(x) + K_S(x)}{\lambda + 1} & \text{if $K_S(x) \neq 0$} \\
      \quad \quad K_{pre}(x) & \text{if $K_S(x) = 0$}
    \end{cases}
\end{equation}
where $K_{pre}$ is the previous k-space before the DC operation. $\lambda$ is the linear combination level, which was 0.05 in our study.

\begin{figure}[htb!]
\centering
\includegraphics[width=0.45\textwidth]{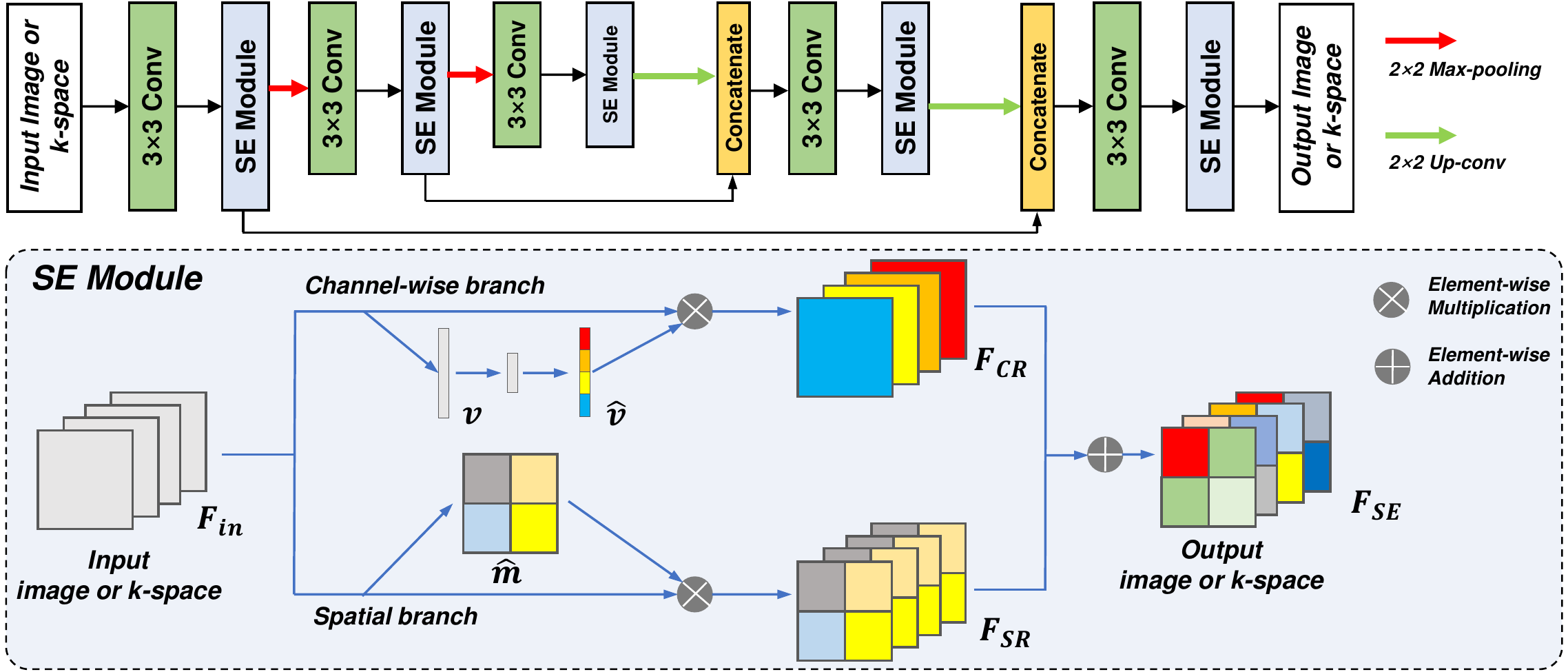}
\caption{The architecture of our proposed SENet. The the SE Module (bottom) in SENet contains two branches for feature recalibration.}
\label{fig:senet}
\vspace{-0.6cm}
\end{figure}

\begin{figure*}[htb!]
\centering
\includegraphics[width=0.88\textwidth]{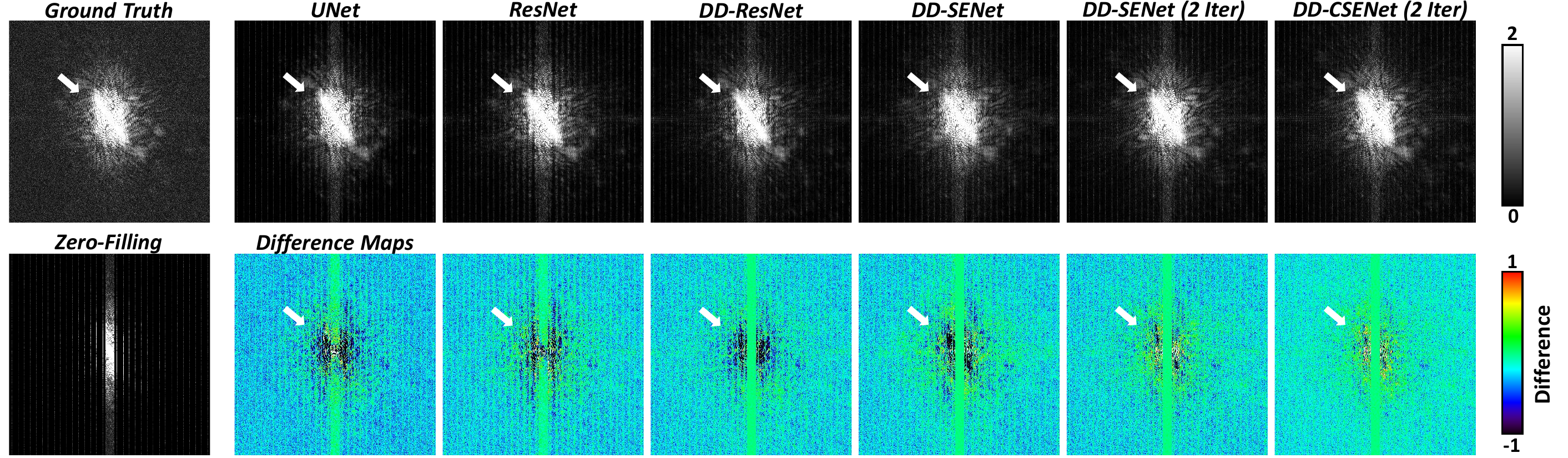}
\caption{Visualizations of the synthetic k-space data of different testing groups. White arrows denote the difference.}
\label{fig:ksapce}
\vspace{-0.5cm}
\end{figure*}

Next, after IFT reconstruction, $K_1$ is input to \textit{I-Net}$_2$ of the 2$^{nd}$ iteration. In the meantime, $I_1$ is element-wise added with $\mathcal{F}^{-1}(K_1)$ using the Cross-Iteration Residual Connection. Thus, the image-domain information of the $1^{st}$ iteration is maintained and passed on to the next iteration, which better extract and incorporate the input features. The output of \textit{I-Net}$_2$ is formulated as:
\begin{equation}
    I_2 = \mathcal{H}_{I_2}(\mathcal{F}^{-1}(K_1) + I_1),
\end{equation}
where $\mathcal{H}_{I_2}$ denotes the SENet in \textit{I-Net}$_2$. Similarly, $K_1$ is also element-wise added with $\mathcal{F}(I_2)$ using the Cross-Iteration Residual Connection to maintain the k-space features of the previous iteration. The output of \textit{K-Net}$_2$ is:
\begin{equation}
    K_2 = \mathcal{P}(\mathcal{H}_{K_2}(\mathcal{P}(\mathcal{F}(I_2) + K1, K_S)), K_S),
\end{equation}
where $\mathcal{H}_{K_2}$ denotes the SENet in \textit{K-Net}$_2$.

Finally, after $N$ iterations, the output $K_N$ is reconstructed into the MRI image output $R_{out}$ using the RSS algorithm.

\subsection{Dual Squeeze-Excitation Network}
Fig.~\ref{fig:senet} presents the structure of SENet that was contained in each \textit{I-Net} or \textit{K-Net} module of DD-CSENet. SENet adopts the contraction-expansion framework. The SE Module in SENet includes two branches for feature recalibration \cite{chen2021ct}. 

In the \textbf{channel-wise branch}, the input image or k-space is formulated as $F_{in}=\left[f_1,f_2,\cdots ,f_C \right]$, where $f_{n}\in R^{H\times W}$ denotes the feature of the $n^{th}$ channel. $H$, $W$, $C$ refers to the height, width, and number of channels of $F_{in}$. $F_{in}$ is first squeezed into a vector $v\in R^C$ with the $n^{th}$ element as $v_n=\frac{1}{H\times W}\sum\nolimits_i^H{\sum\nolimits_j^W{f_n}\left( i,j \right)}$. Then, the vector $v$ is input two fully connected layers and output $\hat{v}=\left[\hat{v}_1,\hat{v}_2,\cdots ,\hat{v}_C \right]$. $\hat{v}$ is then multiplied with $F_{in}$ to recalibrate the channel-wise weights of $F_{in}$, generating $F_{CR}$ as:
\begin{equation}
    F_{CR}=\left[f_1\hat{v}_1,f_2\hat{v}_2,\cdots ,f_C\hat{v}_C \right],
\end{equation}

In the \textbf{spatial branch}, the input image or k-space $F_{in}$ is formulated as $F_{in}=\left[ r^{1,1},\cdots ,r^{i,j},\cdots ,r^{H,W} \right]$, where $r^{i,j}\in R^C$ indicates the feature at the spatial location $\left(i, j\right)$. $F_{in}$ is first squeezed into an area $\hat{m} \in  R^{1 \times H\times W}$ using a $1 \times 1$ convolutional layer. $\hat{m}$ is then multiplied with $F_{in}$ to recalibrate the spatial weights of $F_{in}$, generating $F_{SR}$ as:
\begin{equation}
     F_{SR} = \left[ r^{1,1}\hat{m}^{1,1}, \cdots, r^{i,j}\hat{m}^{i,j}, \cdots, r^{H,W}\hat{m}^{H,W}  \right].
\end{equation}

Next, the two branches are element-wise added to produce the output of the SE Module: $F_{SE} = F_{CR} + F_{SR}$. In this way, SENet can recalibrate both the channel-wise and spatial features of $F_{in}$ to improve the network prediction accuracy.

\subsection{Loss function and implementation details}
The overall end-to-end loss function of DD-CSENet is:
\begin{equation}
    \mathcal{L} = \mathcal{L}_1 + \cdots + \mathcal{L}_m + \cdots + \mathcal{L}_N
\end{equation}
where $\mathcal{L}_m = \lambda_{I_m} l_{I_m}+\lambda_{K_m} l_{K_m}$ represents the total loss of the $m^{th}$ iteration. $\lambda$ represents the loss weights. $l_{I_m}$ or $l_{K_m}$ is the $L_2$ loss between the fully-sampled and predicted images or k-spaces of the $m^{th}$ iteration:
\begin{equation}
    l_{I_m}=\left\| I_m-I_{full} \right\|_2, \quad l_{K_m}=\left\| K_m-K_{full} \right\| _2,
\end{equation}
where $I_{full}$ and $K_{full}$ are fully sampled image and k-space.

The optimal performance of DD-CSENet ($N=2$) was obtained when $\lambda _{I_1}$, $\lambda _{K_1}$, $\lambda _{I_2}$, $\lambda _{K_2}$ were 0.25, 0.25, 1, 1. The DD-CSENet was trained for 400 epochs with a learning rate of $5\times10^{-5}$ and a batch size of 2 using Adam optimizer. The consumed memory was about 12 GB and the training time was about 36 hours on a NVIDIA Tesla V100 graphic card.

\section{Results and Discussion}
\label{sec:result}
\subsection{Predicted synthetic k-space Data}
Multiple methods were tested as listed in Table~\ref{tab:kspace}. The quantification was based on voxel-wise normalized mean-square-error (NMSE), structural similarity (SSIM), and peak signal-to-noise ratio (PSNR). \textit{U-Net} and \textit{ResNet} refer to the k-space methods using U-Net and ResNet. \textit{DD-ResNet} and \textit{DD-SENet} are the dual-domain method using ResNet and SENet. \textit{DD-SENet (2 Iter)} is the iterative dual-domain methods of two-iterations using SENet but without the Cross-Iteration Residual Connection. \textit{DD-CSENet (2 Iter)} is the proposed DD-CSENet method of two-iteration as shown in Fig.~\ref{fig:diagram}.

\begin{table} [htb!]
\setlength{\abovecaptionskip}{0cm} 
\setlength{\belowcaptionskip}{-0.2cm}
\caption{Quantification of the generated k-space data.}
\label{tab:kspace} 
\scriptsize
\centering
\resizebox{0.48\textwidth}{!}{
\begin{tabular}{l  c  c  c  c}
\hline
Methods                     & NMSE$\%$                        & SSIM                        & PSNR                     & P-value$^{\dag}$           \Tstrut\Bstrut\\
\hline 
Zero-Filling                & $19.27 \pm 6.05$                & $0.075 \pm 0.001$           & $54.78 \pm 2.55$         & \textendash                 \Tstrut\Bstrut\\
UNet                        & $10.32 \pm 2.25$                & $0.132 \pm 0.027$           & $57.35 \pm 1.53$         & $< 0.001$                  \Tstrut\Bstrut\\
ResNet                      & $8.06 \pm 2.14$                 & $0.151 \pm 0.026$           & $58.47 \pm 1.96$         & $< 0.001$                  \Tstrut\Bstrut\\
DD-ResNet                   & $7.50 \pm 2.25$                 & $0.183 \pm 0.036$           & $58.19 \pm 2.45$         & $< 0.001$                  \Tstrut\Bstrut\\
DD-SENet                    & $7.08 \pm 2.61$                 & $0.186 \pm 0.021$           & $59.14 \pm 2.35$         & $< 0.001$                  \Tstrut\Bstrut\\
DD-SENet (2 Iter)           & $6.85 \pm 2.26$                 & $0.207 \pm 0.034$           & $59.33 \pm 2.66$         & $0.008$                    \Tstrut\Bstrut\\
DD-CSENet (2 Iter)         & $\mathbf{6.54 \pm 2.56}$                 & $\mathbf{0.221 \pm 0.034}$           & $\mathbf{59.52 \pm 2.41}$         & $< 0.001$                  \Tstrut\Bstrut\\
\hline
\multicolumn{5}{l}{$^{\dag}$P-value of paired t-test on NMSE between the current and previous group.}        \Tstrut\Bstrut\\
\end{tabular}}
\vspace{-0.5cm}
\end{table}

\begin{figure*}[htb!]
\centering
\includegraphics[width=0.88\textwidth]{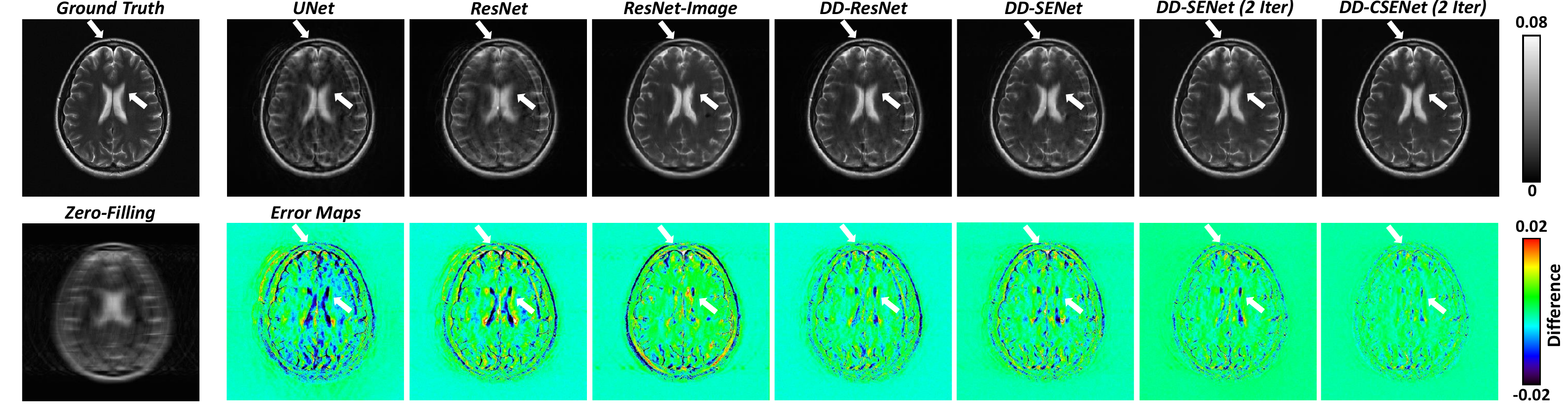}
\caption{Visualizations of the reconstructed MRI images using predicted k-space of different testing groups. White arrows denote the difference.}
\label{fig:recon}
\vspace{-0.4cm}
\end{figure*}

It can be observed that dual-domain methods generally outperformed k-space methods (\textit{DD-ResNet} vs \textit{ResNet}, NMSE 7.50$\%$ vs 8.06$\%$, $p<0.001$). SENet demonstrated better performance than ResNet in the dual-domain framework (\textit{DD-SENet} vs \textit{DD-ResNet}, NMSE 7.08$\%$ vs 7.50$\%$, $p<0.001$). Increasing the number of iterations improved the performance of the dual-domain framework (\textit{DD-SENet (2 Iter)} vs \textit{DD-SENet}, NMSE 6.85$\%$ vs 7.08$\%$, $p=0.008$). Our proposed \textit{DD-CSENet (2 Iter)} using SENet and Cross-Iteration Residual Connection showed higher prediction accuracy than any other testing groups ($p<0.001$). As shown in Fig.~\ref{fig:ksapce}, the synthetic k-space predicted by the proposed \textit{DD-CSENet (2 Iter)} was the most consistent with the ground-truth fully sampled k-space among all testing groups.

\subsection{Reconstructed MRI images}
Table~\ref{tab:recon} lists the quantitative evaluations of the MRI images reconstructed using the synthetic k-space. \textit{ResNet-Image} is the image-domain method using ResNet. It can be observed that image-domain and k-space methods showed quite close performance (\textit{ResNet-Image} vs \textit{ResNet}, NMSE 6.03$\%$ vs 6.12$\%$, $p=0.202$). Our proposed \textit{DD-CSENet (2 Iter)} showed the highest reconstruction accuracy compared to other testing groups ($p<0.001$). As shown in Fig.~\ref{fig:recon}, the reconstructed MRI image by \textit{DD-CSENet (2 Iter)} was the most consistent with the ground-truth MRI images compared to other testing groups.

\begin{table} [htb!]
\setlength{\abovecaptionskip}{0cm} 
\setlength{\belowcaptionskip}{-0.2cm}
\caption{Quantification of the reconstructed MRI images.}
\label{tab:recon} 
\scriptsize
\centering
\resizebox{0.48\textwidth}{!}{
\begin{tabular}{l  c  c  c  c}
\hline
Methods                     & NMSE$\%$                        & SSIM                        & PSNR                     & P-value$^{\dag}$           \Tstrut\Bstrut\\
\hline 
Zero-Filling                & $13.65 \pm 4.21$                & $0.984\pm 0.005$            & $26.09 \pm 1.44$         & \textendash                  \Tstrut\Bstrut\\
UNet                        & $6.97 \pm 1.41$                 & $0.991 \pm 0.003$           & $28.87 \pm 1.56$         & $< 0.001$                  \Tstrut\Bstrut\\
ResNet                      & $6.12 \pm 1.66$                 & $0.991 \pm 0.002$           & $29.50 \pm 1.24$         & $< 0.001$                  \Tstrut\Bstrut\\
ResNet-Image                & $6.03 \pm 1.31$                 & $0.990 \pm 0.001$           & $29.52 \pm 1.12$         & $0.202$                  \Tstrut\Bstrut\\
DD-ResNet                   & $4.05 \pm 0.88$                 & $0.992 \pm 0.001$           & $31.24 \pm 1.10$         & $< 0.001$                  \Tstrut\Bstrut\\
DD-SENet                    & $3.52 \pm 0.88$                 & $0.995 \pm 0.001$           & $31.86 \pm 1.16$         & $< 0.001$                  \Tstrut\Bstrut\\
DD-SENet (2 Iter)           & $2.49 \pm 0.66$                 & $0.997 \pm 0.001$           & $33.39 \pm 1.09$         & $< 0.001$                    \Tstrut\Bstrut\\
DD-CSENet (2 Iter)         & $\mathbf{2.28 \pm 0.57}$        & $\mathbf{0.998 \pm 0.001}$  & $\mathbf{33.76 \pm 1.05}$   & $< 0.001$                  \Tstrut\Bstrut\\
\hline
\multicolumn{5}{l}{$^{\dag}$P-value of paired t-test on NMSE between the current and previous group.}        \Tstrut\Bstrut\\
\end{tabular}}
\vspace{-0.5cm}
\end{table}

\section{Conclusion}
\label{sec:conclusion}
In this paper, we proposed a novel DD-CSENet, consisted of SENet and Cross-Iteration Residual Connection, for the accelerated sparse reconstruction of brain MRI. SENet showed better performance than the commonly-used ResNet under dual-domain frameworks. The Cross-Iteration Residual Connection further improved the network reconstruction accuracy. The proposed DD-CSENet demonstrated state-of-the-art performance in MRI sparse reconstruction, superior to existing image-domain, k-space, and dual-domain methods.  

\section{Compliance with Ethical Standards}
\label{sec:compliance}
The acquisition of the deidentified fastMRI Dataset was conducted retrospectively using human subject data, made available in open access by Center for Advanced Imaging Innovation and Research (CAI$^2$R) in the Department of Radiology at NYU and NYU Langone Health. Curation of these data are part of an IRB approved study.

\section{Acknowledgements}
\label{ssec:acknowlesgement}
This work was funded by Siemens Medical Solutions USA, Inc. The authors would like to express their gratitude to Ian Bowler for his technical support for this project.

\bibliographystyle{IEEEbib}
\bibliography{refs}

\end{document}